\documentclass[10pt,twocolumn,letterpaper]{article}

\usepackage{cvpr}
\usepackage{times}
\usepackage{epsfig}
\usepackage{graphicx}
\usepackage{amsmath}
\usepackage{amssymb}

\usepackage{booktabs}
\usepackage{algorithm}
\usepackage{algorithmicx}
\usepackage{multicol,multirow}
\usepackage{subfigure}
\usepackage[dvipsnames]{xcolor}
\usepackage{pifont}

\usepackage[breaklinks=true,bookmarks=false]{hyperref}

\newcommand\blfootnote[1]{
  \begingroup
  \renewcommand\thefootnote{}\footnote{#1}
  \addtocounter{footnote}{-1}
  \endgroup
}

\newcommand\brfootnote[1]{
  \begingroup
  \renewcommand\thefootnote{}\footnote{#1}
  \addtocounter{footnote}{-1}
  \endgroup
}

\usepackage{hyperref}
\hypersetup{
    colorlinks=true
}

\cvprfinalcopy 

\def\cvprPaperID{5072} 

\begin{document}

\title{TPNet: Trajectory Proposal Network for Motion Prediction}

\author{
   {Liangji Fang}${^{\dag\star}}$ \quad
   {Qinhong Jiang}${^{\dag\star}}$ \quad
   {Jianping Shi}${^{\dag}}$ \quad
   {Bolei Zhou}${^{\ddag}}$\\
   ${^\dag}${SenseTime Group Limited} \qquad
   ${^\ddag}${The Chinese University of Hong Kong}\\
{\tt\small 
\{fangliangji, jiangqinhong, shijianping\}@sensetime.com,}
{\tt\small bzhou@ie.cuhk.edu.hk}
}

\maketitle

\blfootnote{$^{\star}$ indicates equal contribution.}

\begin{abstract}
Making accurate motion prediction of the surrounding traffic agents such as pedestrians, vehicles, and cyclists is crucial for autonomous driving.
Recent data-driven motion prediction methods have attempted to learn to directly regress the exact future position or its distribution from massive amount of trajectory data.
However, it remains difficult for these methods to provide multimodal predictions as well as integrate physical constraints such as traffic rules and movable areas.
In this work we propose a novel two-stage motion prediction framework, Trajectory Proposal Network (TPNet).
TPNet first generates a candidate set of future trajectories as hypothesis proposals, then makes the final predictions by classifying and refining the proposals which meets the physical constraints. By steering the proposal generation process, safe and multimodal predictions are realized. Thus this framework effectively mitigates the complexity of motion prediction problem while ensuring the multimodal output.
Experiments on four large-scale trajectory prediction datasets, \ie the ETH, UCY, Apollo and Argoverse datasets, show that TPNet achieves the state-of-the-art results both quantitatively and qualitatively.$^{1}$ \brfootnote{$^{1}$More information is available at \href{https://decisionforce.github.io/TPNet/}{this link}.}

\end{abstract}

\section{Introduction}

\begin{figure}[t]
\begin{center}
\includegraphics[width=\linewidth]{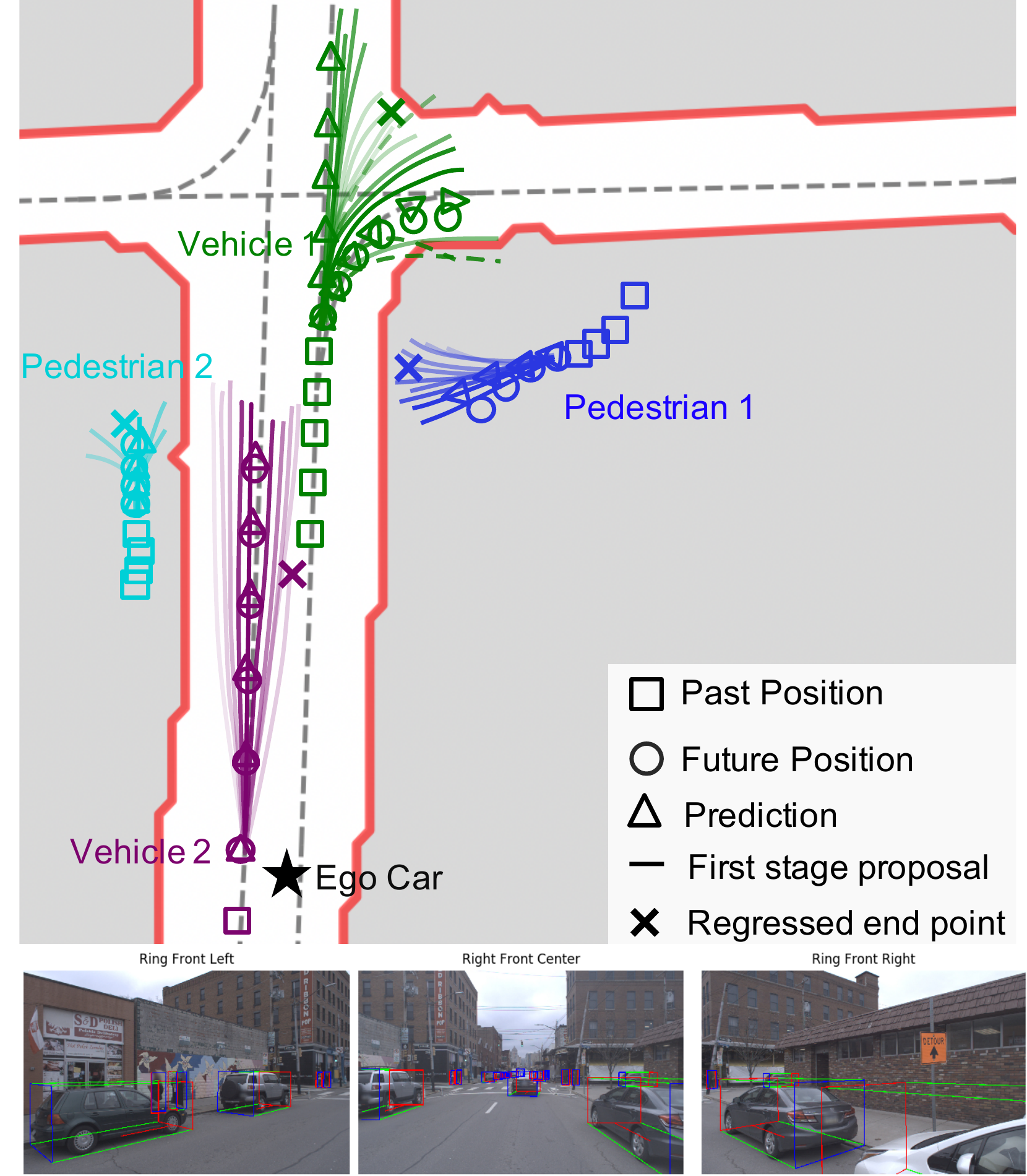}
\end{center}
   \caption{
   The movements of traffic agents are often regularized by the movable areas (white areas for vehicles and gray areas for pedestrians), while there might be multiple plausible future paths for the agents. Thus it requires the motion prediction systems to be able to incorporate the traffic constraints and output multimodal predictions. Our framework generates the predictions with different intentions under physical constraints for both vehicles and pedestrians.
   }
\label{fig:top}
\end{figure}
Predicting the motion of the surrounding traffic agents, such as vehicles, pedestrians, and cyclists, is crucial for the autonomous driving system to make informative and safe decisions.
Traffic agent behaviors tend to be inherently multimodal where there could be multiple plausible intentions for determining their future paths. As illustrated in Fig.~\ref{fig:top}, Vehicle 1 in green could turn right or go straight under this scenario when only a limited number of observations are received.
Moreover the movements of the traffic agents are not only determined by their intentions but also regularized by the nearby traffic rules such as the possible movable areas.
For example, vehicles should drive on the road and pedestrians should walk on sidewalks or crosswalks. Thus reliable motion prediction should involve the modeling of the agent's previous trajectory as well as the traffic constraints for the target. 
Ensuring safe and multimodal predictions is critical for autonomous driving systems.


Early work on motion prediction considers the time-series prediction task by utilizing Kalman Filter based dynamic models \cite{barth2008will, batz2009recognition} or Gaussian mixture models~\cite{deo2018would}, ~\etc. However, these models are sensitive to the observation noise and become unreliable for long-term prediction due to the failure of modeling agent's intention.
Recently, many data-driven motion prediction approaches based on deep neural networks have been developed~\cite{alahi2016social,sgan,hasan2018mx,  lee2017desire,ma2018trafficpredict, sadeghian2018sophie,su2017forecast, varshneya2017human,zhang2019sr}.
Most of them attempt to learn the motion patterns by directly regressing the exact future positions or its distributions from the large amount of trajectory data. The multimodal predictions are generated by sampling from the predicted distribution ~\cite{lee2017desire,zhao2018applying,leung2016distributional}.
However, it is difficult for the data-driven approach to provide reasonable multimodal prediction when the distributions of future positions for different intentions are large (\eg turn left and turn right).
In order to further ensure predictions complied with the traffic rules, the environment information is often encoded as a semantic map then fed into the neural networks~\cite{djuric2018motion,bansal2018chauffeurnet}. However, these end-to-end deep networks lack the safety guarantee to make the output prediction strictly follow the traffic rules or semantic map, while it is difficult for them to effectively incorporate the surrounding physical constraints. 

In this work, we propose a novel two-stage framework called Trajectory Proposal Network (TPNet) to better handle multimodal motion prediction and traffic constraints.
In the first stage, TPNet predicts a rough future end position to reduce the trajectory searching space, and then generates hypothesis as a set of possible future trajectory proposals based on the predicted end point.
In the second stage, TPNet performs classification and refinement on the proposals, then outputs the proposal with highest score as the final prediction.
Proposals with different intentions can be generated in the first stage to realize diverse multimodal predictions.
Prior knowledge such as the movable area constraint is utilized to filtering results of proposals, making this module more effective and transparent.
Extensive experimental results have shown that proposing and refining the future trajectories makes the motion prediction more accurate than the ones which directly regress the future positions.

The contributions of this paper are summarized as follows: 
1) we propose a unified two-stage motion prediction framework for both vehicles and pedestrians. 
2) This framework can incorporate the prior knowledge in the proposal generation process to ensure predictions with multimodal prediction where multiple intentions of the agents are taken into consideration, as well as the compliance of traffic rules and conditions. 
3) We achieve the state-of-the-art results on the recent large-scale trajectory prediction datasets ETH~\cite{pellegrini2009you}, UCY~\cite{lerner2007crowds}, ApolloScape~\cite{ma2019trafficpredict} and Argoverse~\cite{Argoverse}.

\section{Related work}
\label{sec:related}
 
Motion prediction methods can be roughly divided into two categories, the classic methods and the deep learning based methods.
Most of the classic motion based algorithms use kinematics equations to model the agent's motion and predict the future location and the maneuver of the vehicle.
Comprehensive overview of these approaches can be found in ~\cite{lefevre2014survey,rudenko2019human}.
For future location prediction, statistical models such as polynomial fitting~\cite{houenou2013vehicle}, Gaussian processes~\cite{laugier2011probabilistic, tran2014online}, Gaussian mixture models~\cite{deo2018would} have been deployed. Kalman Filter based dynamic models  ~\cite{barth2008will, batz2009recognition} have been also wildly used for motion prediction.
For maneuver recognition, models like Bayesian networks~\cite{schreier2014bayesian}, Hidden Markov models~\cite{deo2018would,laugier2011probabilistic}, SVMs~\cite{aoude2010threat, mandalia2005using}, random forest classifiers~\cite{schlechtriemen2015will} are extensively explored. 
Some of them propose to use scene information to improve prediction~\cite{kitani2012activity,pool2017using}.
These classical methods model the inherent behaviors based only on the previous movements without considering the uncertainty of driver's decision, thus they can not achieve satisfactory performance in long-term prediction.

Recently many deep learning-based methods have been used for motion prediction~\cite{khosroshahi2016surround, kim2017probabilistic,kumar2013learning,yoon2016multilayer}.
Most of them focus on how to extract useful information from the environment.
Convolutional Neural Networks (CNN) Encoder-Decoder is proposed in ~\cite{yagi2018future} to extract features from agents' past positions and directions and directly regress the future positions.
In~\cite{djuric2018motion} the vehicle's location and context information are encoded as binary masks, and a perception RNN is proposed to predict vehicles' location heat-map. 
The typical pipeline for learning-based prediction methods first encodes the input features, then uses CNN or Long Short-Term Memory(LSTM) ~\cite{hochreiter1997long} to extract features and regress the future locations ~\cite{altche2017lstm,lee2017convolution,park2018sequence,xin2018intention}. However, for these data-driven and deep learning-based methods it is difficult to guarantee the safety and the physical constraints of the prediction.
There is another pipeline where the possible trajectory set is first generated based on a lot of motion information (speed, acceleration, angular acceleration, \etc) and then optimize the designed cost function to obtain final prediction ~\cite{houenou2013vehicle}. However this method heavily relies on the accuracy in the physical measurements, high definition map and the quality of the trajectory set. 
Different from ~\cite{houenou2013vehicle}, the proposed TPNet could generate complete proposals only based on trajectory locations. The proposed two-stage pipeline performs further refinement of the proposals which reduces the correlation of the generated proposals and guarantees the diversity of the predictions. Meanwhile, by applying prior knowledge into proposal generation process, our method could take into consideration the physical constraints effectively.


\section{Trajectory Proposal Network}
\label{sec:method}

\begin{figure*}
\begin{center}
\includegraphics[width=\linewidth]{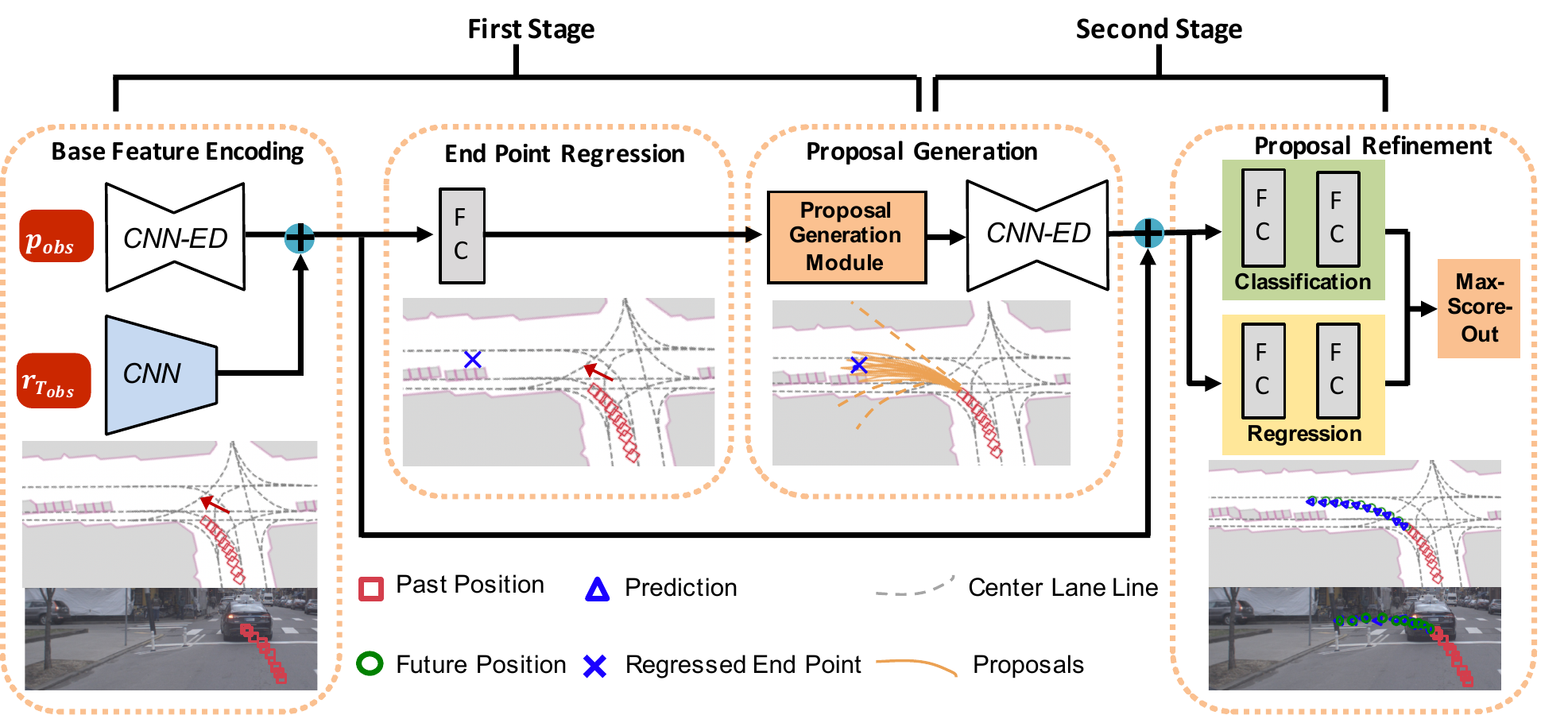}
\end{center}
\caption{Framework of the Trajectory Proposal Network (TPNet).
In the first stage, a rough end point is regressed to reduce the searching space and then proposals are generated. In the second stage, proposals are classified and refined to generate final predictions.
The dotted proposals are the proposals that lie outside of the movable area, which will be further punished.
}
\label{fig:framework}
\end{figure*}



To facilitate the safe and multimodal motion prediction, we propose a novel two-stage framework called Trajectory Proposal Network(TPNet). The framework is shown in Fig.~\ref{fig:framework}:
In the first stage, base features are extracted from the target agent, then a rough end point is predicted to reduce the proposal searching space.
This predicted end point is then utilized to generate proposals.
In the second stage, proposals are classified to find the most possible future trajectory and then are refined to ensure the diversity of final predictions.

By monitoring the proposals generated from the proposal generation process in the first stage, the deep learning based prediction method could be more interpretable and flexible.
Given the generated proposals, the second stage of the TPNet only needs to choose the most plausible trajectory, which simplifies the prediction problem compared to previous methods of directly regressing the trajectory. Furthermore, it is convenient to debug and explain the possible error predictions by examining the outputs from the two stages respectively. 



\subsection{Base Feature Encoding Module}
\label{sec:base_feature}
The \emph{Base Feature Encoding Module} is designed as an encoder-decoder network due to its flexibility of extending different kinds of input features to the module. 
The encoder and decoder blocks consist of several convolutional and deconvolutional layers, respectively. The detailed model structure is illustrated in Fig.~\ref{fig:framework}.

The module takes a series of past positions $p_{obs} = \{p_0, p_1,..., p_{T_{obs}}\}$ in time interval $[0, T_{obs}]$ of the target agent and its surrounding road information $r_{T_{obs}}$ as input, the road information is optional for different dataset.
The road information is represented by many semantic elements, \eg, lane line, cross walks, \etc, and is related to agent's position.
For simplicity, we encode the road information as an image and draw targets past positions onto the image, same as ~\cite{djuric2018motion}.
A small backbone ResNet-18 ~\cite{he2016deep} is used to extract features from the road semantic image.

\subsection{Proposal Generation}
\label{sec:proposal_generation}
In this section, we introduce the detailed process of Proposal Generation.
There are two proposal generation methods depending on whether the road information is utilized or not.
The Base Proposal Generation only uses the position information and can be applied to datasets without road information.
When combined with road information, the multimodal proposal generation can generate proposals for each possible intentions, ensuring a more compact set of hypotheses.

\subsubsection{Problem Definition}
\label{sec:definition}
In our TPNet, we model the agent trajectory in a limited time as a continuous curve to enable efficiency, flexibility, and robustness.
Instead of the traditional representation with discrete point sequence~\cite{ma2018trafficpredict,djuric2018motion} prediction, the  continuous curve~\cite{houenou2013vehicle} avoids inefficient combinatorial explosion of future trajectory set and the lack of physical constraint in some combinations. 
By varying fewer parameters of the curve, we can generate a set of curves flexibly. Curve representation is also robust to noises and could reflect motion tendency and intention. 
We choose polynomial curve to represent the trajectory due to its simplicity ~\cite{houenou2013vehicle}. To find the best polynomial fitting degree, we conduct experiments with different degrees and calculate the fitting errors for trajectories of time length $T = T_{obs} + T_{pre}$, where $T_{obs}$ is the length for the history observations and $T_{pre}$ is the length for the future predictions.
We choose cubic curve with a balance of accuracy and complexity.
The average fitting error is $0.048 \ m$ for pedestrians on ApolloScape dataset, and $0.068 \ m$ for vehicles on Argoverse dataset, which is accurate enough for most cases (detailed analysis can be found in supplemental material).

\begin{figure}
\begin{center}
\includegraphics[width=0.9\linewidth]{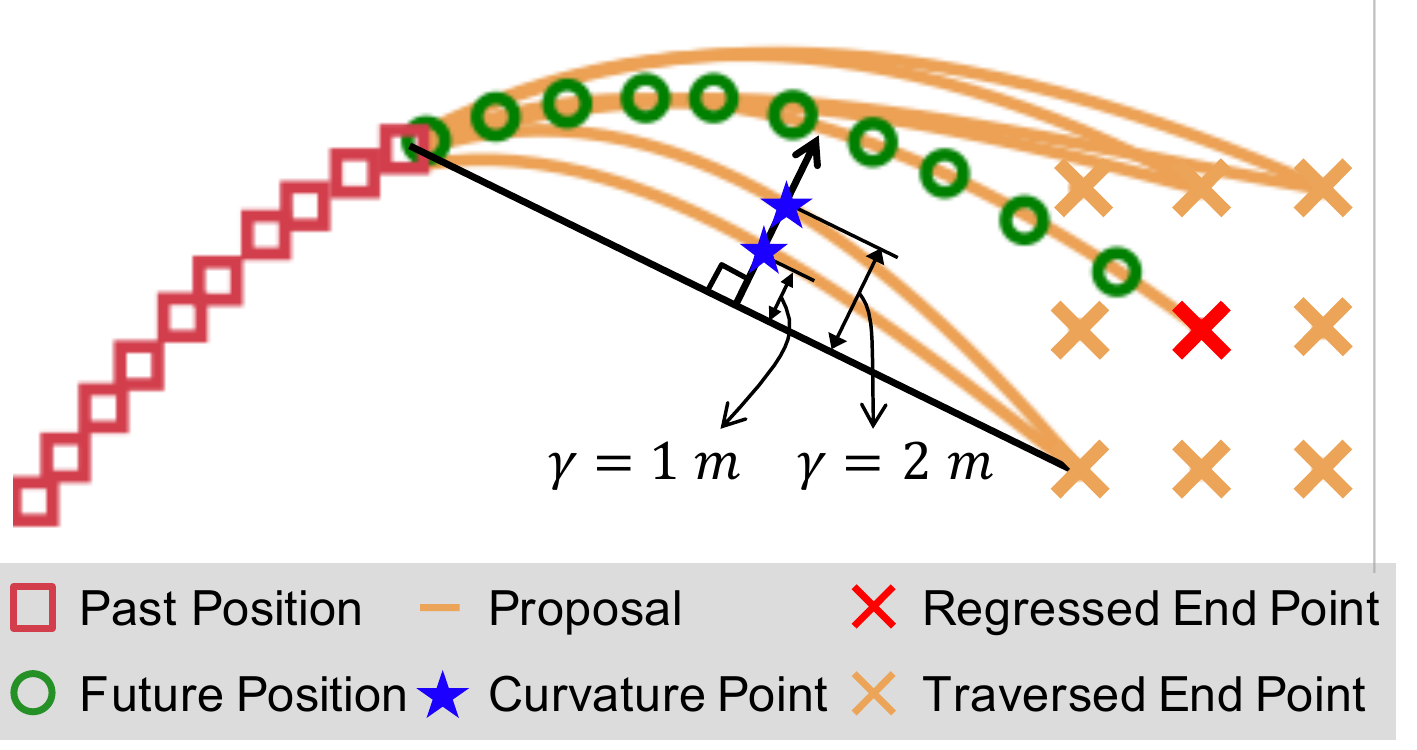}
\end{center}
\caption{Illustration of proposal generation. Proposals are generated around the end point predicted in the first stage. $\gamma$ is used to control the shape of the proposal.}
\label{fig:proposal}
\end{figure}

Since the curve is sensitive to the parameters and difficult to optimize, we propose to use a set of points to represent the curve: two control points, \ie, the end point and curvature point (as shown in Fig.~\ref{fig:proposal}), along with the past points. 
The curvature point reflects curve's crook degree and is determined by a distance variable named $\gamma$. 
$\gamma$ is defined as the distance between the trajectory curve and the mid-point of the current point and end point as shown in Fig. \ref{fig:proposal}.
Encoding curvature point as $\gamma$ allows to generate curves flexibly with different crook degrees.

\subsubsection{Base Proposal Generation}
\label{sec:basic_proposal_generation}

A good proposal generation process should have the ability of generating complete proposals based on less trajectory information.
Hence the Base Proposal Generation method generates proposals only based on the trajectory positions, which are the one of the most fundamental and common features provided by almost all trajectory prediction datasets~\cite{Argoverse,lerner2007crowds,ma2019trafficpredict,pellegrini2009you}.
Given the past positions of an agent, proposals could be generated by varying different control points under curve representation defined in Sec.~\ref{sec:definition}.
Based on the end point $p_{e}$ predicted in the first stage, possible end points can be generated by enumerating a $N \times N$ grid centered at $p_{e}$:
$$p_{ep} = \{(x_e +  interval * i, y_e + interval * j)\}_{i,j \in [-N/2, N/2]},$$
where $p_{ep}$ is the possible end points set, $(x_e, y_e)$ is the coordinate of $p_{e}$, $interval$ and $N$ are the interval and size of the grid.
By varying the values of $\gamma$, different curvature points for each possible end point could be generated.


\begin{figure}
\begin{center}
\includegraphics[width=0.8\linewidth]{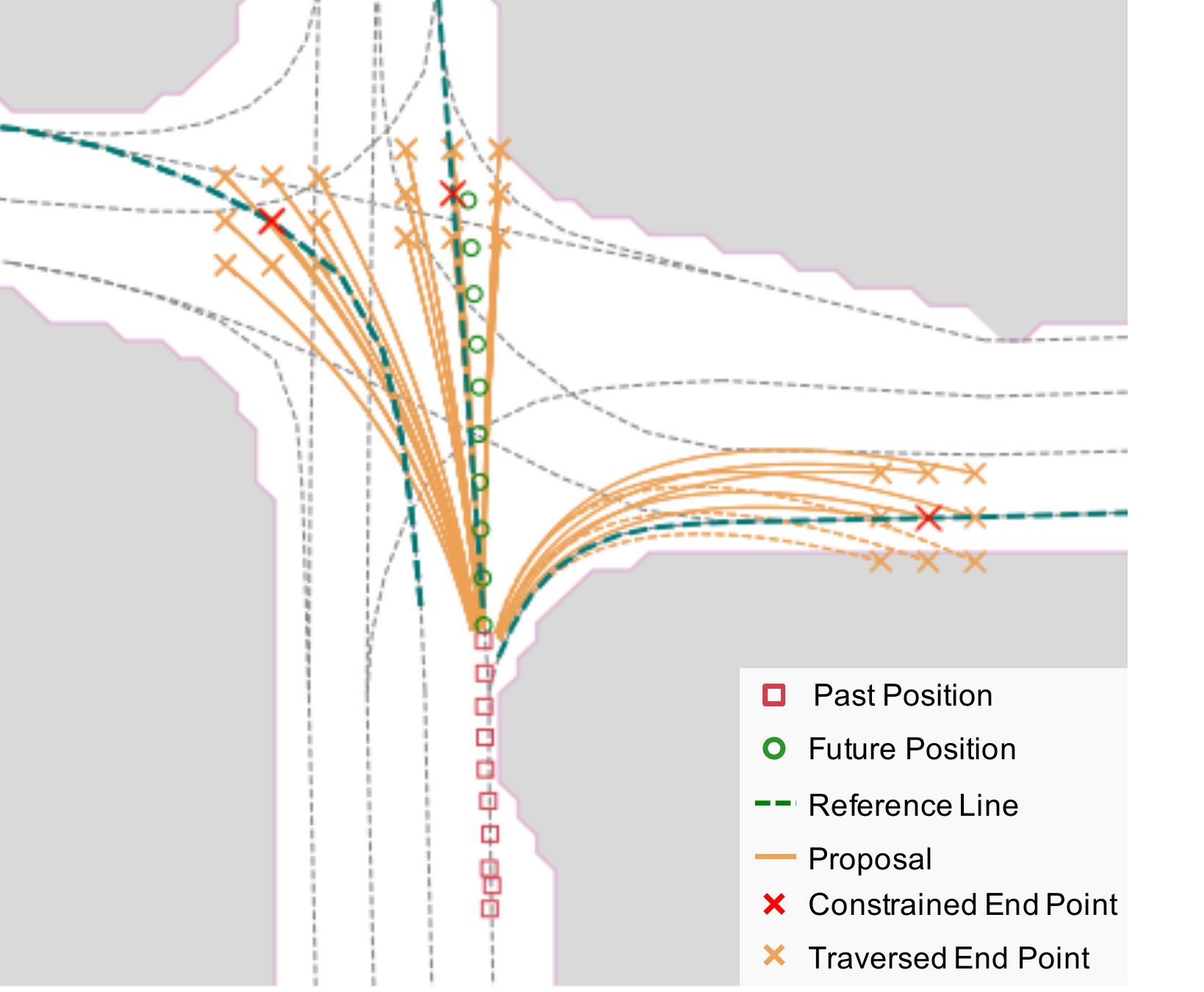}
\end{center}
\caption{Illustration of multimodal proposal generation using road information. The reference lines indicate the possible center lane lines that the vehicle could dive in. Best viewed in color.}
\label{fig:proposal_mm}
\end{figure}

Finally, proposals are generated only based on positions using Eq.~\ref{eq:prposals}.
\begin{equation}
proposals = \{f(p_{obs}, p_{ep}^{'}, \gamma)\},
\label{eq:prposals}
\end{equation}
where $f(\cdot)$ is the cubic polynomial fitting function, $p_{ep}^{'} \in p_{ep}$ and $\gamma \in [-2,-1,0,1,2]$.

\subsubsection{Multimodal Proposal Generation}
\label{sec:mm_proposal_generation}
Base Proposal Generation has strong dependency on the regressed end point in the first stage which might lead to low diversity of the generated proposals.
Multimodal Proposal Generation takes use of road information to generate multiple end points since road has strong constraints on vehicles.
Based on the basic elements of the road information (lane lines and their directions, ~\etc) and vehicle's past positions, we can obtain a set of reference lines represent the possible center lane lines the vehicle will reach ~\cite{Argoverse}. Hence Eq.~\ref{eq:prposals} could be extended to generate multiple proposal sets for different reference lines.

Specifically, the relative 1D end position displacement $d_{ep}$ along the reference line is predicted rather than the 2D end point $p_{e}$. Then we sample the future end point on each reference line based on the predicted $d_{ep}$ which reduce the dependency on the single regressed end point and ensures the diversity of predictions.
Lastly the proposals are generated for each sampled end point using Eq.~\ref{eq:prposals}. The process is illustrated in Fig.~\ref{fig:proposal_mm}.

\subsection{Proposal Classification and Refinement}
\label{sec:class_refine}
Given a set of proposals, the \emph{Classification Module} chooses the best proposal while the \emph{Refinement Module} refines the end points and $\gamma$ of the proposals.

\textbf{Classification Module.}
During training, binary class label, denotes good trajectory or not, is assigned to each proposal.
We define the average distance between the uniformly sampled points of ground truth and proposal trajectory curves as the criterion of proposal's quality, noted as:
\begin{equation}
AD = \frac{1}{N} \sum_{i=1}^{N} \| p_{gt}^{i} - p_{pp}^{i}\|,
\label{eq:ad}
\end{equation}
where $N$ is the number of sampled points, $p_{gt}^{i}$ and $p_{pp}^{i}$ are the $i$-$th$ sampled point of ground truth trajectory and proposal, respectively.
We assign positive label to a proposal that has an $AD$ lower than a threshold, \eg $1 m$.
The remaining proposals are potential negative samples.
To avoid the overwhelming impact of too many negative samples, we adopt uniform sampling method to maintain the ratio between the negatives and positives as 3:1.

\textbf{Refinement Module.} 
For proposals refinement, we adopt the parameterization of the 2 coordinates and 1 variate:
\begin{equation}
\left\{
\begin{array}{lll}
t_x = x_e^{gt} - x_e^{pp},\\
t_y = y_e^{gt} - y_e^{pp}, \\
t_{\gamma} = \gamma^{gt} - \gamma^{pp},
\end{array} \right.
\label{eq:modification}
\end{equation}
 where $(x_e^{gt}, y_e^{gt})$ and $(x_e^{pp}, y_e^{pp})$ are the end point coordinates of ground-truth trajectory and proposal. $t_x$, $t_y$ and $t_{\gamma}$ are the supervised information used during training.

\textbf{Model Designing.}
For each proposal, we use the same encoder-decoder module mentioned in Sec.~\ref{sec:base_feature} to extract features. Then the base features are concatenated with the proposal features. Last two fully connected layers are utilized to classify and refine the proposals, respectively.

\subsection{Prior Knowledge}
\label{sec:prior}
Prior knowledge such as vehicles tend to drive on the road will make the trajectory prediction results more stable and safe.
However, DNN-based solutions cannot guarantee these constraints due to the complexity and unexplained nature of the model.

Thanks to the proposal based pipeline, we can use the prior knowledge to filter the proposals explicitly.
Combined with historical trajectory and high-definition maps, the polygonal area where the agent can travel in the future is determined, namely movable area. 
We propose to explicitly constrain the predicted trajectories during inference by decaying the classification scores of the proposals outside of the movable area using Eq.~\ref{eq:decay}:
\begin{equation}
score = score * e^{\frac{-r^2}{\sigma^2}}
\label{eq:decay}
\end{equation}
where $r$ is the ratio that proposal trajectory points outside of the movable area, and $\sigma$ is the decaying factor.

Compared to abandoning the prediction results outside of the movable area, decaying the classification scores ensures the diversity of the predictions.


\subsection{Objective Function}
\label{sec:objective_function}
During training, we minimize a multi-task loss as:
\begin{equation}
L = L_{ep}(p_{e}, p_{e}^*) + \frac{1}{N}\sum_i L_{cls}(c_i, c_i^*) + \frac{\alpha \sum_i L_{ref}(t_i, t_i^*)}{N_{pos} + \beta N_{neg}},
\label{eq:loss}
\end{equation}
where $p_{e}$ and $p_{e}^*$ are the predicted end point and corresponding ground-truth, $c_i$ and $t_i$ are the predicted confidence and trajectory parameters for each proposal, $c_i^*$ and $t_i^*$ are the corresponding ground-truth labels, $\alpha$ is the weight term. 
Euclidean loss is employed as the end point prediction loss $L_{ep}$ and the refinement loss $L_{ref}$.
Binary cross entropy loss is employed as the classification loss $L_{cls}$.
Due to the multimodal property of the future trajectory, we use positive samples along with part of randomly sampled negative samples to calculate the refinement loss and a $\beta$ to control the ratio of sampled negatives.

\section{Experiments}
\label{sec:exp}

\begin{table*}[tb!]
\begin{center}
\setlength{\tabcolsep}{1.2mm}{
 \begin{tabular}{c|c|c|c|c|c|c|c|c|cc}
  \textbf{Metric} & \textbf{Dataset}   & \textbf{S-LSTM}~\cite{alahi2016social} & \textbf{S-GAN}~\cite{gupta2018social} & \textbf{Liang}~\cite{liang2019peeking} & \textbf{Li}~\cite{li2019way} & \textbf{SoPhie}~\cite{sadeghian2019sophie} & \textbf{STGAT}~\cite{huang2019stgat}  & \textbf{TPNet-1} & \textbf{TPNet-20} \\
  \hline
  \hline
    \multirow{5}*{ADE}  & \textbf{ETH}      & 0.73 / 1.09          & 0.61 / 0.81                & - / 0.73                & - / \textbf{0.59}                & - / 0.70                & 0.56 / 0.65    & 0.72 / 1.00     & \textbf{0.54} / 0.84 \\
    ~                   & \textbf{HOTEL}    & 0.49 / 0.79          & 0.48 / 0.72                & - / 0.30                & - / 0.46                & - / 0.76                & 0.27 / 0.35             & 0.26 / 0.31     &  \textbf{0.19} / \textbf{0.24} \\
    ~                   & \textbf{UNIV}     & 0.41 / 0.67          & 0.36 / 0.60                & - / 0.60                & - / 0.51                & - / 0.54                & 0.32 / 0.52             & 0.34 / 0.55     & \textbf{0.24} / \textbf{0.42} \\
    ~                   & \textbf{ZARA1}    & 0.27 / 0.47          & 0.21 / 0.34                & - / 0.38                & - / \textbf{0.22}                & - / 0.30                & 0.21 / 0.34             & 0.26 / 0.46     & \textbf{0.19} / 0.33 \\
    ~                   & \textbf{ZARA2}    & 0.33 / 0.56          & 0.27 / 0.42                & - / 0.31                & - / \textbf{0.23}                & - / 0.38                & 0.20 / 0.29             & 0.21 / 0.33     & \textbf{0.16} / 0.26 \\
  \hline
    \textbf{AVG}    &                       & 0.45 / 0.72          & 0.39 / 0.58                & - / 0.46                & - / \textbf{0.40 }               & - / 0.54                & 0.31 / 0.43             & 0.36 / 0.53     & \textbf{0.27} / \textbf{0.42} \\
  \hline
  \hline
    \multirow{5}*{FDE} & \textbf{ETH}       & 1.48 / 2.35           & 1.22 / 1.52             & - / 1.65             & - / 1.30             & - / 1.43             & \textbf{1.10} / \textbf{1.12}& 1.39 / 2.01          & 1.12 / 1.73 \\
    ~                  & \textbf{HOTEL}     & 1.01 / 1.76           & 0.95 / 1.61             & - / 0.59             & - / 0.83             & - / 1.67             & 0.50 / 0.66          & 0.48 / 0.58          &  \textbf{0.37} / \textbf{0.46} \\
    ~                  & \textbf{UNIV}      & 0.84 / 1.40           & 0.75 / 1.26             & - / 1.27             & - / 1.27             & - / 1.24             & 0.66 / 1.10          & 0.68 / 1.15          & \textbf{0.53} / \textbf{0.94} \\
    ~                  & \textbf{ZARA1}     & 0.56 / 1.00           & 0.42 / 0.69             & - / 0.81             & - / \textbf{0.49}             & - / 0.63             & 0.42 / 0.69          & 0.55 / 0.99          & \textbf{0.41} / 0.75 \\
    ~                  & \textbf{ZARA2}     & 0.70 / 1.17           & 0.54 / 0.84             & - / 0.68             & - / \textbf{0.55}             & - / 0.78             & 0.40 / 0.60          & 0.43 / 0.72          & \textbf{0.36} / 0.60 \\
  \hline
    \textbf{AVG} &                          & 0.91 / 1.52           & 0.78 / 1.18             & - / 1.00             & - / 0.89             & - / 1.15             & 0.62 / \textbf{0.83}          & 0.71 / 1.08          & \textbf{0.56} / 0.90 \\
  \hline
 \end{tabular}}
 \caption{Comparison with baseline methods on ETH and UCY benchmark for $T_{pre}=8$ and $T_{pre}=12$ (8 / 12). Each row represents a dataset and each column represents a method. 20V-20 means that use variety loss and sample 20 times during test time according to \cite{gupta2018social, huang2019stgat}. TPNet-20 means we chose the best prediction from proposals with top-20 classification scores.}
\label{tab:eth}
\end{center}
\end{table*}

\begin{table}
\begin{center}
\setlength{\tabcolsep}{1.2mm}{
 \begin{tabular}{c|c|c|c|c|cc}
  \textbf{Metric} & \textbf{Type}   & \textbf{S-LSTM} & \textbf{S-GAN} & \textbf{StarNet}~\cite{zhu2019starnet} & \textbf{TPNet} \\
  \hline
  \hline
    \multirow{3}*{ADE}  & \textbf{Ped}  & 1.29                & 1.33    & 0.79     & \textbf{0.74} \\
    ~                   & \textbf{Veh}  & 2.95                & 3.15    & 2.39     &  \textbf{2.21} \\
    ~                   & \textbf{Cyc}  & 2.53                & 2.53    & 1.86     & \textbf{1.85} \\
  \hline
    \multicolumn{2}{c|}{WSADE}         & 1.89                & 1.96    & 1.34     & \textbf{1.28} \\
  \hline
  \hline
    \multirow{3}*{FDE} & \textbf{Ped}  & 2.32             & 2.45      & 1.52       &   \textbf{1.41} \\
    ~                  & \textbf{Veh}  & 5.28             & 5.66       & 4.29      &  \textbf{3.86} \\
    ~                  & \textbf{Cyc}  & 4.54             & 4.72      & 3.46       & \textbf{3.40} \\
  \hline
    \multicolumn{2}{c|}{WSFDE}         & 3.40             & 3.59      & 2.50          & \textbf{2.34\footnotemark[2]} \\
  \hline
 \end{tabular}}
\caption{Comparison with other methods on the ApolloScape dataset. In the table, Veh, Ped and Cyc indicate agent types of Vehicle, Pedestrian and Cyclist, respectively. Since the ground-truth labels of test set are not released, we only report the unimodal result of S-GAN and TPNet.}
\label{tab:apollo} 
\vspace{-20pt}
\end{center}
\end{table}
TPNet is evaluated on four public datasets, ETH~\cite{pellegrini2009you}, UCY~\cite{lerner2007crowds}, ApolloScape \cite{ma2018trafficpredict} and Argoverse \cite{Argoverse}.
\textbf{ETH} and \textbf{UCY} datasets focus on the pedestrian trajectory prediction. Totally there are five subsets, named ETH, HOTEL, ZARA-01, ZARA-02 and UCY. We follow the same data
preprocessing strategy as Social GAN ~\cite{gupta2018social}. There are two settings for the length of trajectories, $T_{obs}=T_{pre}=3.2s$ and $T_{obs} = 3.2s, T_{pre} = 4.8s$. The time interval is set as $0.4s$ for both settings, which results in 8 frames for observation and 8/12 frames for prediction.
\textbf{ApolloScape} contains bird eye view coordinates of target agents' trajectories along with the trajectories of their surrounding agents. There are three object types need to be predicted, namely vehicle, pedestrian, cyclist.
For the length of trajectories, ApolloScape set $T_{obs} = T_{pre} = 3s$ and time interval as $0.5s$, which results in 6 frames for both observation and prediction.
\textbf{Argoverse} dataset focuses on the prediction of vehicle trajectories. Besides the bird eye view coordinates of each vehicle, Argoverse dataset also provides the high-definition maps. 
For the length of trajectories, Argoverse set $T_{obs} = 2s, T_{pre} = 3s$ and time interval as $0.1s$.
The training, validation and testing sets contain 205942, 39472 and 78143 sequences respectively.

\textbf{Evaluation Metrics.}
Average Displacement Error (ADE) and Final Displacement Error (FDE) are the most used metrics in motion prediction.
ApolloScape also uses the weighted sum of ADE (WSADE) and weighted sum of FDE (WSFDE) as metrics among different agents types.
Argoverse also calculates minimum ADE (minADE), minimum FDE (minFDE) and Drivable Area Compliance (DAC).

\begin{itemize}
	
%
	\item \emph{WSADE/WSFDE}: weighted sum of ADE/FDE among different agents types.
	
	\item \emph{minADE/minFDE}: is the minimum ADE/FDE among multiple predictions (up to K=6) .
	
	\item \emph{DAC}: is the ratio of the predicted positions inside the drivable area.
    
\end{itemize}

\footnotetext[2]{The number on \href{https://openaccess.thecvf.com/content_CVPR_2020/papers/Fang_TPNet_Trajectory_Proposal_Network_for_Motion_Prediction_CVPR_2020_paper.pdf}{CVPR website} is wrong. We correct this value in current release.}

\textbf{Baselines.} 
Since the multimodal proposal generation and safety guarantee in our proposed method are dependent on high-definition maps, the comparison methods are divided into two groups.
The first group consists of methods do not use high-definition maps, including Social LSTM~\cite{alahi2016social} and Social GAN~\cite{gupta2018social}. These baselines are compared on ApolloScape, ETH and UCY dataset.
The second group consists of methods that use high-definition maps, including Nearest Neighbor~\cite{Argoverse} and LSTM ED~\cite{Argoverse}. These baselines are compared on Argoverse dataset.
\begin{itemize}
	\item Social LSTM (S-LSTM): uses LSTM to extract features of trajectory and propose social pooling to model social influence for pedestrian trajectory prediction. 
	\item Social GAN (S-GAN): proposes a conditional GAN which takes the trajectories of all agents as input.
	\item Nearest Neighbor (NN): weighted Nearest Neighbor regression using top-K hypothesized centerlines. 
	\item LSTM ED: LSTM Encoder-Decoder model with road map information as input.
\end{itemize}

\textbf{Implementation Details.}
For network input, road elements within $70 m \times 70 m$ relative to the target agent is encoded into a semantic map with resolution of 0.5 m/pixel. 
ResNet-18 ~\cite{he2016deep} is used to extract features of the semantic map.
During training, we use data augmentation by randomly rotating and flipping the trajectories.
The ratio between negative and positive samples is set to 3:1 and positive $AD$ threshold is set to $3 m$ experimentally.
We optimize the network using Adam~\cite{kingma2014adam} with batch size of 128 for 50 epochs, and learning rate 0.001 with a decay rate 0.9.

\subsection{Comparison with Baselines}
\label{sec:exp_comparison}
The effectiveness of the proposed two-stage framework is evaluated on ETH, UCY and Apollo dataset with only target's bird eye view past positions as input in Tab.~\ref{tab:eth} and Tab.~\ref{tab:apollo}.
To validate the multimodal prediction and safety guarantee of our proposed method, experiments are conducted on Argoverse dataset as shown in Tab.~\ref{tab:argo}. 

\textbf{Evaluation of Two-stage Framework.}
The proposed TPNet is compared with the baselines on ETH and UCY datasets in terms of two metrics ADE and FDE in Tab.~\ref{tab:eth}. 
Following the evaluation methods in S-GAN, we report the results as TPNet-1 and TPNet-20, where TPNet-1 is the prediction with highest classification score while the TPNet-20 result is the best prediction among the predictions with \emph{top-K} classification scores. The results show that the TPNet-1 result already outperforms Social LSTM and the multi-modal results of Social GAN. After using the TPNet-20 result, TPNet is competitive with all baselines on all datasets.
Note that TPNet only uses the past positions of the target agent while other baselines also utilizing the positions of around agents, which could potentially make our method worse on some datasets.

Then, the performance results of TPNet and the comparison methods on ApolloScape dataset are shown in Tab.~\ref{tab:apollo}. 
From the table we can see that TPNet outperforms the baseline methods on all agent types. 
Specifically, TPNet performs better on vehicle trajectory prediction and we believe it is because that the curve representation is more friendly to vehicle trajectories.


\begin{table}[]
\begin{center}
\setlength{\tabcolsep}{1.4mm}{
\begin{tabular}{@{}c|ccccc@{}}
Methods & ADE & FDE & minADE & minFDE  & DAC \\ 
\hline
\hline
NN ~\cite{chang2019argoverse}           &  3.45             &  7.88             & 1.71     & 3.29     & 0.87  \\
LSTM ED ~\cite{chang2019argoverse}      &  2.96             &  6.81             &  2.34             & 5.44              & 0.90  \\
TPNet                                   &  2.33             &  5.29             &  2.08             & 4.69              & 0.91  \\ 
TPNet-map                               &  2.23    &  4.71             &  2.04             & 4.23              & 0.96  \\
TPNet-map-safe                          &  2.23             &  4.70             &  2.03             & 4.22              & \textbf{0.99}  \\
TPNet-map-mm                            &  \textbf{2.23}             &  \textbf{4.70}             &  \textbf{1.61}             & \textbf{3.28}              & 0.96  \\ 
\hline
\end{tabular}}
\end{center}
\caption{Comparison with baseline methods on the Argoverse test set.}
\label{tab:argo}
\end{table}

\textbf{Evaluation of Multimodal Prediction.}
TPNet-map-mm in Tab.~\ref{tab:argo} generates proposals with different intentions based on reference lines mentioned in Sec.~\ref{sec:proposal_generation}.
In the table, TPNet is referred as our method with only past positions as input, TPNet-map as our method with past positions and road semantic map as input. TPNet-map-safe and TPNet-map-mm are referred as using prior knowledge to constrain the proposals and generating multimodal proposals, respectively.
In order to evaluate the diversity of the prediction method, Argoverse~\cite{Argoverse} uses minADE and minFDE as metrics. These two metrics calculate the best ADE and FDE among \emph{K} number of samples for each target trajectory.
After the proposals with different intentions are generated, minADE and minFDE are improved by 60cm and 1m, respectively.
Furthermore, the proposed TPNet could generate multimodal predictions even without the use of reference lines. As shown in Tab.~\ref{tab:eth}, the TPNet-20 results on ETH and UCY dataset outperforms the TPNet-1 result by a large margin without the use of reference lines.
Because of the proposal generation process, predictions with different intentions could be ensured more effective.

\textbf{Evaluation of Safety Guarantee.}
To evaluate the effectiveness of safety guarantee mentioned in Sec.~\ref{sec:prior}, we show the experiment results on Argoverse dataset in Tab.~\ref{tab:argo}.
Tab.~\ref{tab:argo} shows that TPNet outperforms the baselines proposed in Argoverse ~\cite{Argoverse} by a large margin, especially on FDE. This indicates that TPNet could generate more accurate end point.

Furthermore, after taking road semantic map as input, TPNet-map achieves a better results.
However the prediction results still may outside the drivable area as the DAC metric still has the room for improvements.

By decaying the classification scores of proposals outside the drivable area using Eq.~\ref{eq:decay}, DAC is improved to 0.99 for TPNet-map-safe which indicates our proposed method could generate more safe prediction results.



\subsection{Ablation Study}
\label{sec:exp_ablation}
In this section, we will illustrate the effectiveness of each part of TPNet. We choose Argoverse dataset to do the ablation study for two reasons, 1) the scale of Argoverse dataset is larger than others, 2) Argoverse dataset provides the ground-truth labels for the validation set.

\textbf{Two-stage Framework.}
To further validate the effectiveness of modeling trajectory prediction as a two-stage framework, experiments on removing the classification and regression modules step by step are conducted. The results are shown in Tab.~\ref{tab:stage}.
By removing the classification and regression simultaneously, the model achieves 4.01 m on FDE metric. The predicted trajectory is obtained by sampling the positions on the curve fitted by the past positions and predicted end position.
Then a cascade regressor is utilized to refine the predicted end point and it further improves the FDE by 5 cm as shown in the second row in Tab.~\ref{tab:stage}.
Finally the complete two-stage pipeline is experimented and the FDE could be further improved by 8 cm.

\textbf{Grid Size.}
The proposed method relies on the quality of generated proposals. The influence of grid size for proposals generation is shown in Tab.\ref{tab:grid}. TPNet will have better results when grid range is set to $6m \times 6m$. As the grid range grows, the performance becomes worse as the searching space becomes larger. And smaller interval size is better.


\begin{table}[]
\begin{center}
\begin{tabular}{@{}cc|cc@{}}
Regression & Classification & ADE & FDE \\
\hline
\hline
\ding{55}  &   \ding{55}    & 2.00 & 4.01 \\
\checkmark &   \ding{55}    & 1.85 & 3.96 \\
\checkmark &   \checkmark   & \textbf{1.75} & \textbf{3.88} \\
\hline
\end{tabular}
\end{center}
\caption{Ablation Study on the effectiveness of different stages on the Argoverse validation dataset.}
\label{tab:stage}
\end{table}

\begin{figure*}
\begin{center}
\includegraphics[width = 1\linewidth]{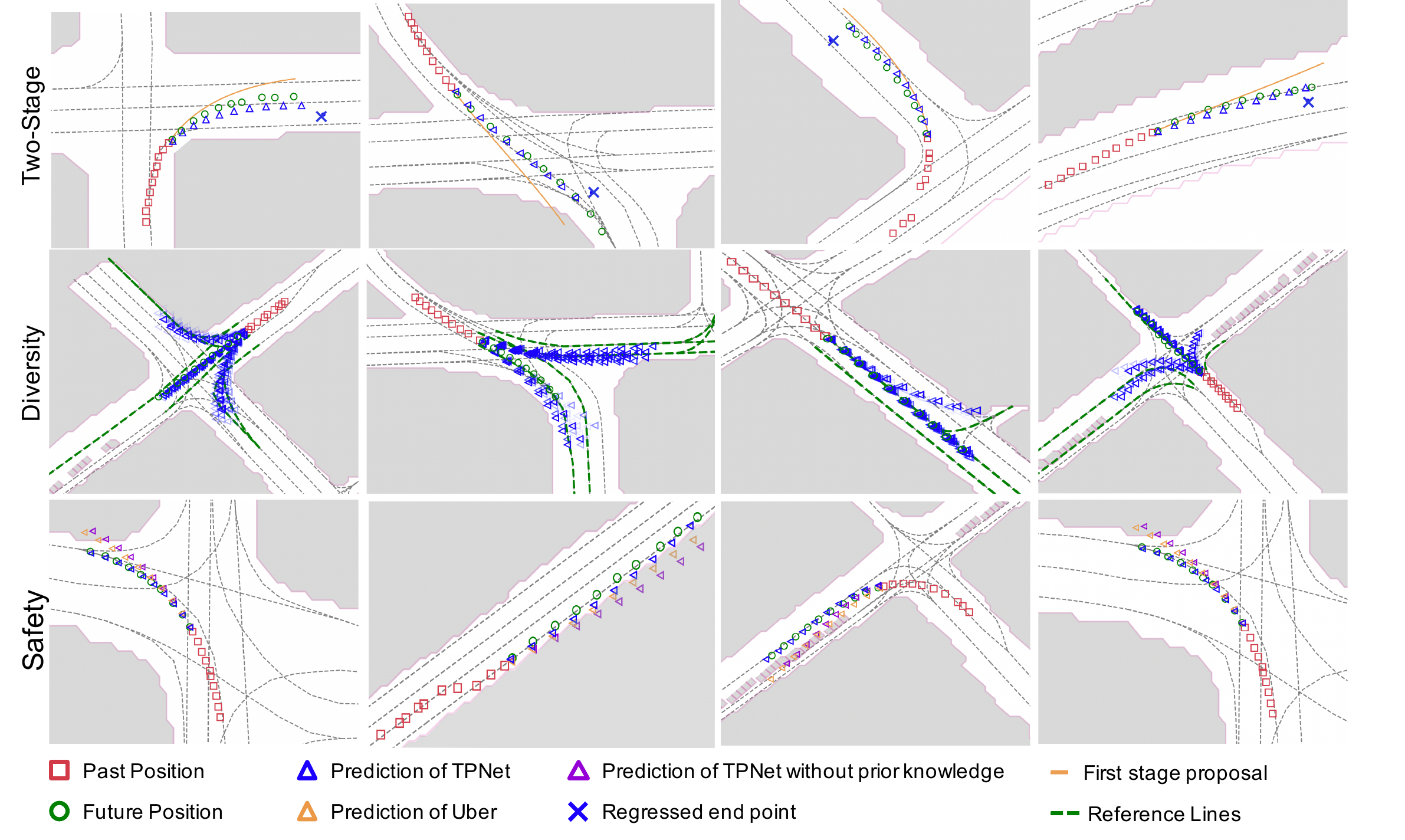}
\end{center}
   \caption{Qualitative results on the effectiveness of each components of TPNet on Argoverse dataset. From top to bottom rows illustrate the effectiveness on the two-stage framework, diversity and safety, respectively. Better viewed in color.}
\label{fig:argo_qualitative}
\end{figure*}





\begin{table}[]
\begin{center}
\begin{tabular}{@{}ccc|cc@{}}
Range (m)   & Interval (m)  & \#Anchor & ADE & FDE \\
\hline
\hline
$6\times 6$ &   1            & 245   & \textbf{1.75}    & \textbf{3.87}	\\
$6\times 6$ &   1.5          & 125   & 1.78             & 3.89  \\
$6\times 6$ &   3            & 45    & 1.84             & 4.01  \\
$10\times10$&   1.67         & 245   & \textbf{1.75}    & 3.88  \\
$10\times10$&   2.5          & 125   & 1.76             & 3.88  \\
$10\times10$&     5          & 45    & 1.84             & 4.01  \\
$20\times20$&   3.3          & 245   & 1.77             & 3.93  \\
$20\times20$&   5            & 125   & 1.79             & 3.98  \\
\hline
\end{tabular}
\end{center}
\caption{Ablation study on the impact of different grid size for anchor generation on the Argoverse validation dataset.}
\label{tab:grid}
\vspace{-15pt}
\end{table}

\subsection{Qualitative Evaluation}
\label{sec:exp_qualitative}
Predicting the motion of traffic agent is challenge because the agent may have different intentions under the same scenario. 
Furthermore, the possible future paths are not only determined by their intentions but also constrained by the nearby traffic rules.
The qualitative results on Argoverse validation set are shown in Fig.~\ref{fig:argo_qualitative}. Most of the selected scenarios are nearing crossroad. Fig.~\ref{fig:argo_qualitative} shows that our method could generate more safe and diverse predictions.

\textbf{Two-stage Framework.}
The effectiveness of the proposed two-stage framework is shown in the first row of Fig.~\ref{fig:argo_qualitative}. The regressed end point might be inaccurate, however the classification and regression processes will refine the prediction results.

\textbf{Multimodal Output.}
In the second row of Fig.~\ref{fig:argo_qualitative}, prediction results under scenarios nearing the crossroad are shown. We can observe multimodal predictions around each possible intentions. Furthermore, the predictions of each intention are also diverse, for example, a vehicle might follow the center lane line or deviate the center lane line. 

\textbf{Safety.}
In the last row of Fig.~\ref{fig:argo_qualitative}, we show the results of TPNet (purple triangle), Uber~\cite{djuric2018motion} (yellow triangle) and TPNet with safety-guaranteed (blue triangle). Uber~\cite{djuric2018motion} encodes the road elements into a raster image and use CNN to regress the future positions. As can be seen in the figure, input the semantic road map to DNN could not ensure the safety of prediction while the proposed decaying function Eq.~\ref{eq:decay} is more reliable.
\section{Conclusion}
In this work we propose a two-stage pipeline for more effective motion prediction. The proposed two-stage TPNet first generates the possible future trajectories served as proposals and uses a DNN based model to classify and refine the proposals.
Multimodal predictions are realized by generating proposals for different intentions.
Furthermore, safe prediction can also be ensured by filtering proposals outside the movable area. 
Experiments on the public datasets demonstrate the effectiveness of our proposed framework.
The proposed two-stage pipeline is flexible to encode prior knowledge into the deep learning method. For example, we can use lamp status which indicates the intention of vehicles to filter the proposals, which will be included in the future work.


{\small
\bibliographystyle{ieee_fullname}
\bibliography{egbib}
}

\end{document}